\documentclass[runningheads]{llncs}
\usepackage{graphicx}
\usepackage{booktabs} % for professional tables
\usepackage{xcolor}
\usepackage{multirow}
\usepackage{amsmath}
\usepackage{mathrsfs}
\usepackage{bm}
\usepackage{amssymb}
\usepackage{hyperref}
\hypersetup{
    colorlinks=true,
    linkcolor=magenta,
    filecolor=magenta,      
    urlcolor=magenta,
    pdftitle={Overleaf Example},
    pdfpagemode=FullScreen,
    }
\usepackage{tensor}
\usepackage{caption}
\usepackage{subcaption}

\newcommand{\myparagraph}[1]{\vspace{0pt}\noindent{\bf{#1}}}

\begin{document}
\title{Uncertainty-Guided Progressive GANs for Medical Image Translation}
%
%\titlerunning{Abbreviated paper title}
% If the paper title is too long for the running head, you can set
% an abbreviated paper title here
%
\author{
Uddeshya Upadhyay\inst{1} \and
Yanbei Chen\inst{1} \and
Tobias Hepp\inst{1,2} \and \\
Sergios Gatidis\inst{1,2} \and
Zeynep Akata\inst{1,2}
}
%
% index{Upadhyay, Uddeshya}
% index{Chen, Yanbei}
% index{Hepp, Tobias}
% index{Gatidis, Sergios}
% index{Akata, Zeynep}
\authorrunning{U. Upadhyay et al.}
% First names are abbreviated in the running head.
% If there are more than two authors, 'et al.' is used.
%
\institute{
University of T{\"u}bingen \and
Max Planck Institute for Intelligent Systems
}
\maketitle              % typeset the header of the contribution
\begin{abstract}
Image-to-image translation plays a vital role in tackling various medical imaging tasks such as attenuation correction, motion correction, undersampled reconstruction, and denoising. Generative adversarial networks have been shown to achieve the state-of-the-art in generating high fidelity images for these tasks.
However, the state-of-the-art GAN-based frameworks do not estimate the uncertainty in the predictions made by the network that is essential for making informed medical decisions and subsequent revision by medical experts and has recently been shown to improve the performance and interpretability of the model.
In this work, we propose an uncertainty-guided progressive learning scheme for image-to-image translation. By incorporating aleatoric uncertainty as attention maps for GANs trained in a progressive manner, we generate images of increasing fidelity progressively. We demonstrate the efficacy of our model on three challenging medical image translation tasks, including PET to CT translation, undersampled MRI reconstruction, and MRI motion artefact correction. Our model generalizes well in three different tasks and improves performance over state of the art under full-supervision and weak-supervision with limited data. Code is released here:
\href{https://github.com/ExplainableML/UncerGuidedI2I}{https://github.com/ExplainableML/UncerGuidedI2I}
\end{abstract}

%\vspace{-25pt}
\section{Introduction}
%\vspace{-8pt}
In the medical domain, each imaging modality reflects particular physical properties of the tissue under examination. This results in images with different dimensionality, spatial resolution, and contrast. Various imaging modalities provide a complimentary stream of information for clinical diagnostics or technical pre and post-processing steps. Moreover, acquiring medical images is susceptible to various kinds of noise and modality-specific artefacts. To remedy these issues, translating images between different domains is of great importance. 

Inter-modal image-to-image translation can potentially replace additional acquisition procedures, reducing examination costs and time. Besides, intra-modality image-to-image translation enables complex artefact and noise correction.
For example, attenuation correction of positron emission tomography (PET) data is challenging in situations where no density distribution is available from computed tomography (CT) data, as in the case for stand-alone PET scanners or combined PET/magnetic resonance imaging (MRI). In these situations, the generation of pseudo-CTs from PET data can be helpful. Further examples are related to image reconstruction and/or correction in MRI: Reconstruction of undisturbed artifact-free images is hard to achieve with traditional methods; deep-learning-based image-to-image translation can solve this challenge.
In particular, generative adversarial networks (GAN) based on convolutional neural networks (CNN) have proven to provide a high visual quality of the generated synthetic images. However, predictions of GANs can be unreliable, and particularly in medical applications, the quantification of uncertainty is of high importance for the interpretation of the results.
In this work, we propose a generic end-to-end model that introduces high-capacity conditional progressive GANs to synthesize high-quality images, using aleatoric uncertainty estimates as the guide to focus on improving image quality in regions where the network is highly uncertain about the prediction. 
We perform experiments on three challenging and vital medical imaging tasks: PET to CT translation, undersampled MRI reconstruction, and motion correction in MRI. Moreover, we empirically demonstrate the efficacy of our model under weak supervision with limited data.

%\vspace{-7pt}
\section{Related Works}
%\vspace{-8pt}
Traditional machine learning techniques for medical image translation rely on explicit feature representations~\cite{huynh2015estimating,zhong2016predict,kustner2017mr,rueda2013single}. More recently, convolutional neural networks have been proposed for various image translation tasks \cite{litjens2017survey,shin2016deep,dou2016multilevel,havaei2017brain,kamnitsas2017efficient,chen2017low} and state-of-the-art performance is achieved by generative adversarial networks \cite{nie2018medical,wolterink2017deep,yang2018low,zhang2019image,pix2pix,yi2019generative,progan,medgan,armanious2019unsupervised,upadhyay2019mixed,upadhyay2019robust}. 
The existing methods propose conditional GAN architectures with deterministic outputs that typically uses $\mathcal{L}_1/\mathcal{L}_2$-based fidelity loss for the generator assumes a pixel-wise \textit{homoscedasticity} and also assumes the pixel-wise error (i.e., residual) to be \textit{independent and identically distributed} (i.i.d) following a Laplace or Gaussian distribution. This is a limiting assumption as explained in~\cite{whatuncer,uncercyc,wang2019aleatoric}.
While these methods can provide synthetic images of high visual quality, the image content may still deviate significantly from the corresponding ground-truth. This results in overconfidence or misinterpretation with negative consequences, particularly in the medical domain. 
There have been recent works on quantifying aleatoric and epistemic uncertainty in task-specific medical imaging algorithms like classification, segmentation, super-resolution etc~\cite{nair2020exploring,wang2019aleatoric,wang2018automatic,wang2019automatic,tanno2017bayesian}
quantifying it for general image-to-image translation problem largely remains unexplored.
Thus, the central motivation of our work is to provide measures of uncertainty for image-to-image translation tasks that can contribute to safe applications of results.

Moreover, recent work has shown that high-capacity generators that are progressive in nature lead to high-quality results as described in~\cite{medgan,armanious2019unsupervised,progan}. However, the progressive generation of high-quality images remains unguided
without specifically attending to poorly translated regions.
Prior works indicate a correlation between estimated uncertainty and prediction error~\cite{zhang2019reducing,tanno2017bayesian,uncercyc}. 
We exploit this relationship for the progressive enhancement of synthetic images, which has not been investigated by prior work before.

%\vspace{-7pt}
\section{Uncertainty-Guided Progressive GAN (UP-GAN)}
%\vspace{-8pt}

Let $A$ and $B$ be two image domains with a set of images
$S_{A} := \{a_1, a_2 ... a_n\}$ and $S_{B} := \{b_1, b_2 ... b_m\}$ where $a_i$ and $b_i$ represent the $i^{th}$ image from domain $A$ and $B$ respectively.
Let each image drawn from an underlying \textit{unknown} probability distribution $\mathcal{P}_{AB}$, i.e., $(a_i, b_i) \sim \mathcal{P}_{AB} \forall i$ have $K$ pixels, and $u_{ik}$ represent the $k^{th}$ pixel of a particular image $u_i$.
Our goal is to learn a mapping from domain $A$ to $B$ ($A \rightarrow B$) in a paired manner, i.e., learning the underlying conditional distribution $\mathcal{P}_{B|A}$ from the set of given samples $\{(a_i, b_i)\}$, following the distribution $\mathcal{P}_{AB}$.
For a given image $a_i$ in domain $A$, the estimated image in domain $B$ is called $\hat{b}_i$. The pixel wise error is defined as $\epsilon_{ij} = \hat{b}_{ij} - b_{ij}$.
While the existing framework models the residual as the i.i.d as described above, we relax that assumption by modelling the residual as non i.i.d
variables and learning the optimal distribution from the dataset, as described in the following.

\begin{figure}[t]
\centering
\includegraphics[width=\textwidth]{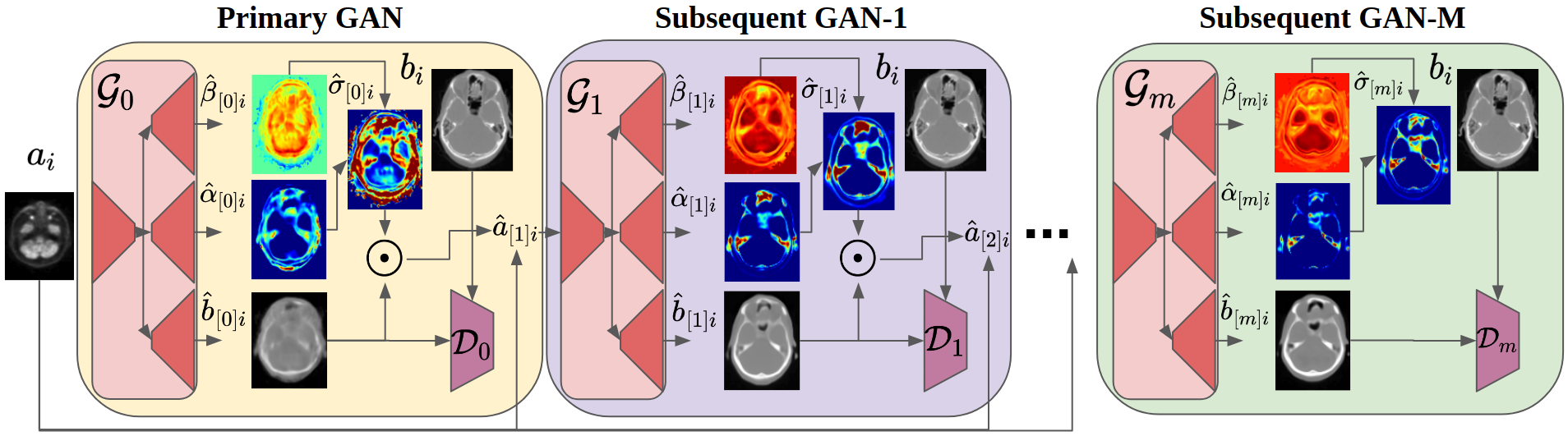}
%\vspace{-15pt}
\caption{
Uncertainty-guided Progressive GANs (UP-GAN): The primary GAN takes the input image from domain $A$, while subsequent GANs absorb outputs from the preceding GAN (see Eq.~\ref{eq:eq3} and \ref{eq:eq4}). %UP-GAN progressively generates high fidelity images in domain $B$. 
Explicitly guided by the attention maps, the uncertainty maps are estimated from the preceding GAN.
}
\label{fig:arch}
%\vspace{-8pt}
\end{figure}

Figure~\ref{fig:arch} shows our model that consists of cascaded GANs, where each generator is capable of estimating the aleatoric uncertainty, along with generating images. 
Our solution alleviates the aforementioned %above described
limitations of recent methods by modelling the underlying per-pixel residual distribution as \textit{independent} but \textit{non-identically} distributed \textit{zero-mean generalized Gaussian distribution} (GGD) as in~\cite{uncercyc}, where the network learns to predict the optimal \textit{scale} ($\alpha$) and \textit{shape} ($\beta$) of the GGD for every pixel, 
Therefore, $\hat{b}_{ij} = b_{ij} + \epsilon_{ij}$ with, 
$
    \epsilon_{ij} \sim GGD(\epsilon; 0, \alpha_{ij}, \beta_{ij}) \equiv \beta_{ij}(2\alpha_{ij}\Gamma(\beta^{-1}_{ij}))^{-1} \exp{\left(-\alpha_{ij}^{-1}|\epsilon|^{\beta_{ij}}\right)}
$.
We generate images in multiple phases, with each phase generating output images along with the aleatoric uncertainty estimates. The outputs from one phase serve as the input to the subsequent GAN in the next phase, explicitly guided by the attention map derived from uncertainty estimates. Importantly, this uncertainty-based guidance enforces the model to focus on refining the uncertain regions that are likely to be poorly synthesized, resulting in progressively improving quality.

Our framework is composed of a sequence of $M$ GANs, where the $m^{th}$ GAN is represented by a pair of networks, generator and discriminator, given by, $(\mathcal{G}_m(\cdot; \theta_m), \mathcal{D}_m(\cdot; \phi_m))$. Both the generator and discriminator can have arbitrary network architecture as long as generator can estimate \textit{aleatoric uncertainty} as described in~\cite{uncercyc}. We choose all the discriminators to be the patch discriminators from~\cite{pix2pix} and generators to be modified U-Net~\cite{ronneberger2015u}, where the head is split into three to estimate the parameters of the GGD as shown in Figure~\ref{fig:arch} and in~\cite{uncercyc}. %GANs are trained progressively as follows.

\textbf{Primary GAN.} We train the first GAN ($\mathcal{G}_0$) using the dataset $S_A$ and $S_B$.
The predictions of the generator are given by $(\hat{\alpha}_{[0]i},\hat{\beta}_{[0]i},\hat{b}_{[0]i})$. The network is trained
with an adaptive fidelity loss function $\mathcal{L}^G_{\alpha\beta}$~\cite{uncercyc} and an adversarial loss $\mathcal{L}^G_{\text{adv}}$~\cite{cycleGAN}, combined as $\mathcal{L}^G_{\text{tot}}$ for the generator ($\mathcal{G}_0(\cdot; \theta_0): A \rightarrow B$):%, 
\begin{small}
\begin{gather}
% \begin{align}
    \mathcal{L}^G_{\alpha\beta}(\hat{b}_{[0]i}, \hat{\alpha}_{[0]i}, \hat{\beta}_{[0]i}, b_{i}) = 
    \frac{1}{K}\bm\sum_{j}  \left(\frac{|\hat{b}_{[0]ij}-b_{ij}|}{\hat{\alpha}_{[0]ij}} \right)^{\hat{\beta}_{[0]ij}} - 
    \log\frac{\hat{\beta}_{[0]ij}}{\hat{\alpha}_{[0]ij}} + \log\Gamma(\hat{\beta}_{[0]ij}^{-1})
    \label{eq:eq1}
    \\
    \mathcal{L}_{\text{adv}}^G = \mathcal{L}_2(\mathcal{D}_1(\hat{b}_{[0]i}),1) \text{ and } 
    \mathcal{L}^G_{\text{tot}} = \lambda_1 \mathcal{L}^G_{\alpha\beta} + \lambda_2 \mathcal{L}_{\text{adv}}^G .
    \label{eq:eq2}
% \end{align}
\end{gather}
\end{small}
The patch discriminator ($\mathcal{D}_1$) is trained using the adversarial loss from~\cite{cycleGAN} given by $\mathcal{L}^D_{\text{adv}} = \mathcal{L}_2(\mathcal{D}^A(b_i),1) + \mathcal{L}_2(\mathcal{D}^A(\hat{b}_{[0]i}),0)$. 

\textbf{Subsequent GANs.} The $m^{th}$ GAN (where $m > 0$) takes the output produced by the $(m-1)^{th}$ GAN, i.e.
$(\hat{\alpha}_{[m-1]i},\hat{\beta}_{[m-1]i},\hat{b}_{[m-1]i})$, along with the original sample $a_i$ from domain $A$ as its input and generates a refined output.
The image estimated by the $(m-1)^{th}$ GAN along with its uncertainty map learns to create the input feature
$f_{[m]i}$ for the $m^{th}$ GAN, where the uncertainty map serves as an attention mechanism to highlight the uncertain regions in the image.
The input $a_{[m]i}$ for the $m^{th}$ generator is given by concatenating $a_i$ and $f_{[m]i}$, i.e.,
\begin{gather}
    \hat{\sigma}_{[m-1]i} = \hat{\alpha}_{[m-1]i}\sqrt{
    \frac{\Gamma(3/\hat{\beta}_{[m-1]i})}
    {\Gamma(1/\hat{\beta}_{[m-1]i})}} 
    \text{, and } 
    f_{[m]i} = \hat{b}_{[m-1]i} \odot \frac{\hat{\sigma}_{[m-1]i}}
    {\bm{\sum}_j \hat{\sigma}_{[m-1]ij}}
    \label{eq:eq3}
    \\
    a_{[m]i} = \mathtt{concat}(f_{[m]i}, a_i)
    \label{eq:eq4}
\end{gather}
The input $a_{[m]i}$ 
for the $m^{th}$ GAN encourages the generator to further refine the highly uncertain regions in the image given the original input context. The generator and the discriminator are trained using $\mathcal{L}^G_{\text{tot}}$ and $\mathcal{L}^D_{\text{adv}}$, respectively.

\textbf{Progressive training scheme.} 
We initialize the parameters $\theta \cup \phi$ sequentially. 
First, we initialize $\theta_1 \cup \phi_1$ using the training set $(S_A, S_B)$
to minimize the loss function given by $\mathcal{L}^G_{\text{tot}}$ and $\mathcal{L}^D_{\text{adv}}$. Then, for the subsequent GANs, we initialize the $\theta_m \cup \phi_m$ ($m>1$) by fixing the weights of all the previous generators and training the $m^{th}$ GAN alone (see Eq.~\ref{eq:eq3} and \ref{eq:eq4} with losses $\mathcal{L}^G_{\text{tot}}$ and $\mathcal{L}^D_{\text{adv}}$). 
Once all the parameters have been initialized (i.e., $\theta_m \cup \phi_m \forall m$), we do further fine tuning by training all the networks end-to-end by combining the loss functions of all the intermediate phases and a significantly smaller learning-rate.  

%\vspace{-7pt}
\section{Experiments}
%\vspace{-10pt}
In this section, we first detail the experimental setup and comparative methods in Section \ref{sec:setup}, and present the corresponding results in Section \ref{sec:results}.

%\vspace{-10pt}
\subsection{Experimental Setup}
\label{sec:setup}
\textbf{Tasks and datasets.}
We evaluate our method on the following three tasks.

\begin{figure}[t]
\includegraphics[width=\textwidth]{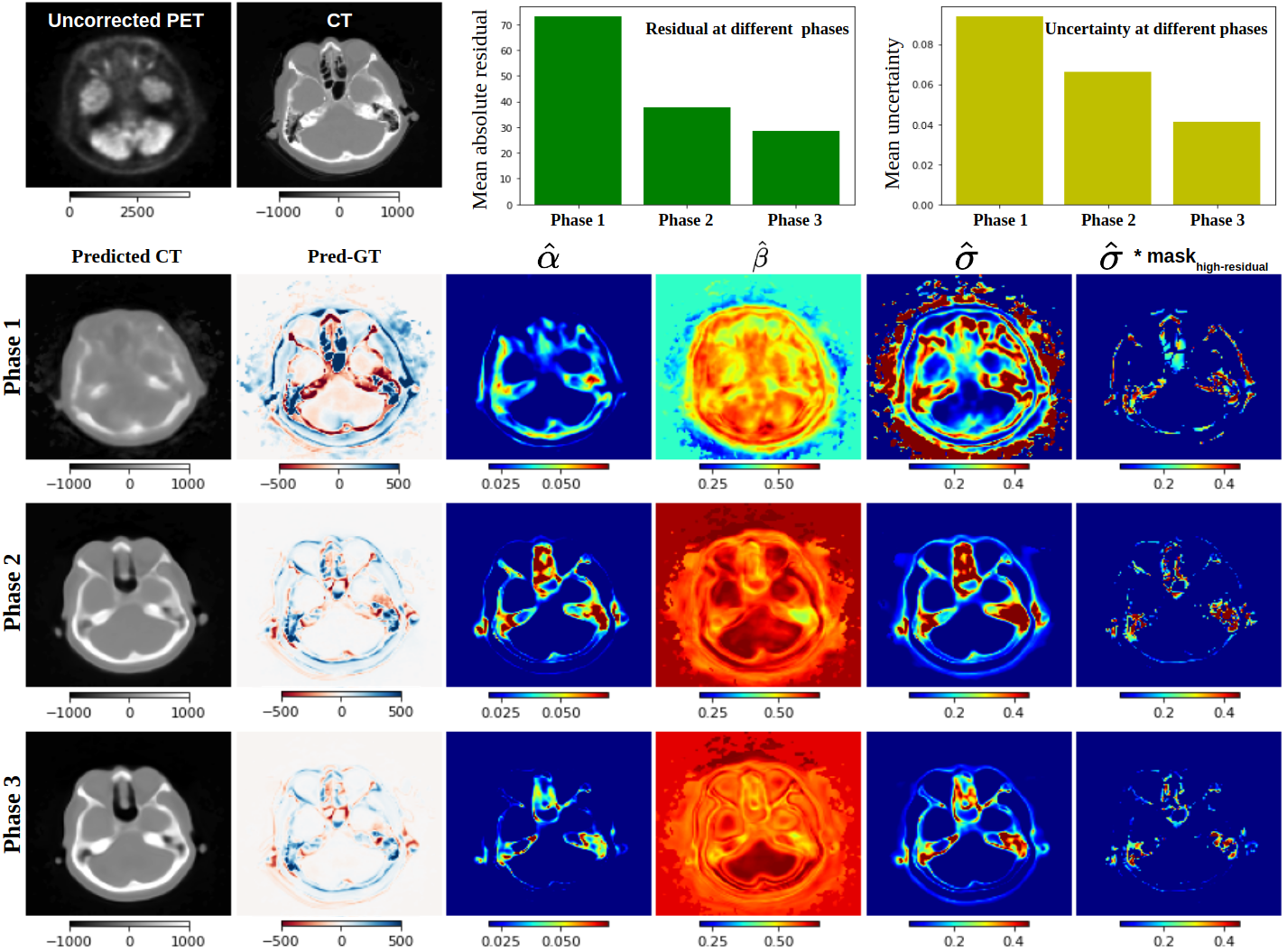}
%\vspace{-15pt}
\caption{Outputs from different phases of UP-GAN (with M=3). (Top) The input (uncorrected PET), the corresponding ground-truth CT, mean residual values over different phases, mean uncertainty values over different phases. (Bottom) Each row shows the predicted output, the residual between the prediction and the ground-truth, the predicted scale ($\alpha$) map, the predicted shape ($\beta$) map, the uncertainty map, and the uncertainty in high residual regions.
}
\label{fig:uncerprog}
%\vspace{-10pt}
\end{figure}

(i) PET to CT translation: We synthesize CT images from 
PET scans to be used for the attenuation correction, e.g. for PET-only scanners or PET/MRI.
We use paired data sets of non-attenuation-corrected PET and the corresponding CT of the head region of 49 patients acquired on a state-of-the-art PET/CT scanner (Siemens Biograph mCT), approved by ethics committee of the Medical Faculty of the University of T{\"u}bingen.
Data is split into 29/5/15 for training/val/test sets. 
Figure~\ref{fig:uncerprog} shows exemplary slices for co-registered PET and CT. 

(ii) Undersampled MRI reconstruction: We translate undersampled MRI images to fully-sampled MRI images.
% where 
We use MRI scans from the open-sourced IXI~\footnote{from \href{https://brain-development.org/ixi-dataset/}{https://brain-development.org/ixi-dataset/}} dataset that consists of T1-weighted (T1w) MRI scans. We use a cohort of 500 patients split into 200/100/200 for training/val/test, and retrospectively create the undersampled MRI with an acceleration factor of $12.5\times$, i.e., we preserve only $8\%$ of the fully-sampled k-space measurement (from the central region) to obtain the undersampled image.

\begin{figure}[t]
\centering
    \includegraphics[width=\textwidth]{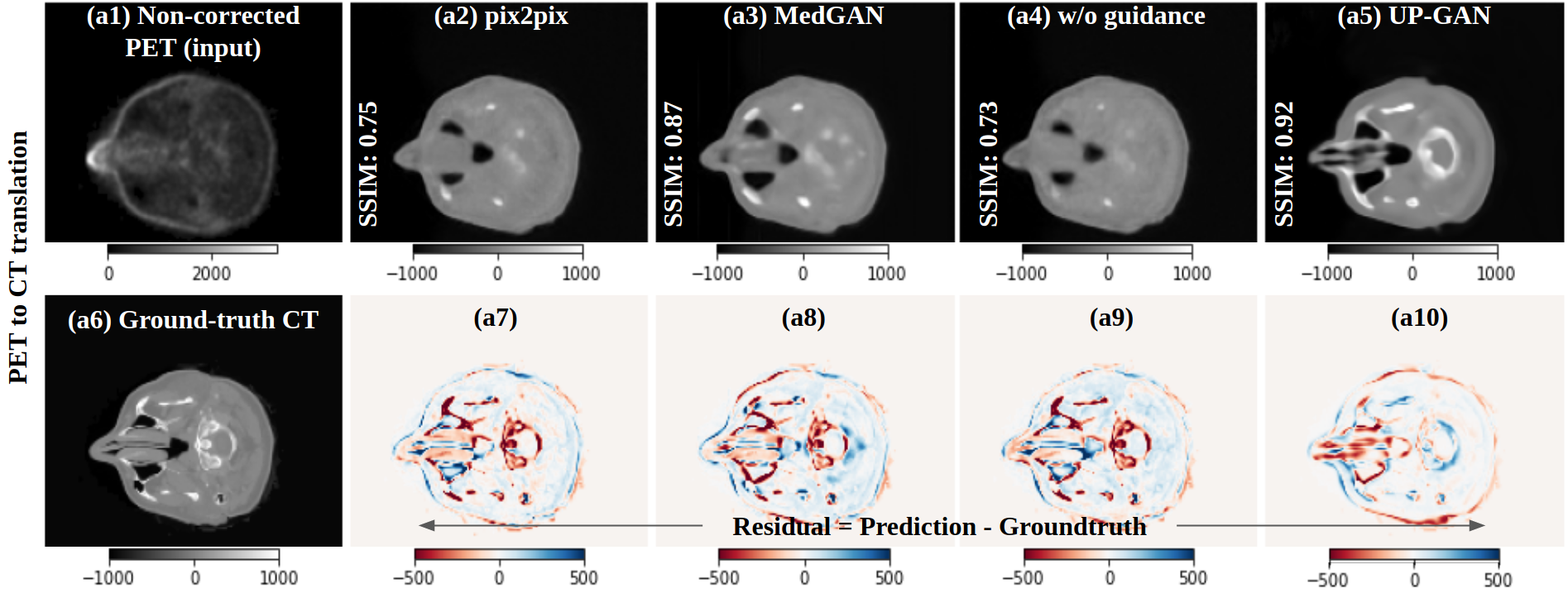}\\
    \vspace{2mm}
    \includegraphics[width=\textwidth]{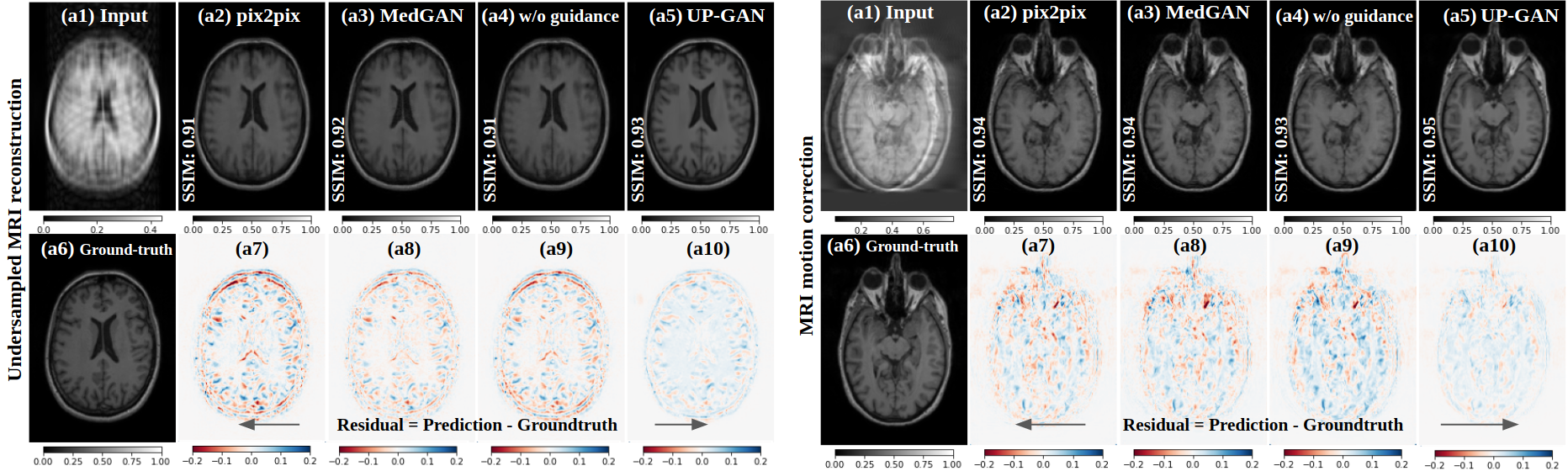}
\caption{Qualitative results. (Top) PET to CT translation. (Bottom) Undersampled MRI reconstruction (left), and MRI motion correction (right).} 
\label{fig:qual_petct_mri}
\end{figure}

(iii) MRI Motion correction: We generate sharp images from motion corrupted images.
% where 
We retrospectively create the motion artefacts in the T1w MRI from IXI following the transformations in the \textit{k-space} as described in~\cite{shaw2019mri}. 
Figure~\ref{fig:qual_petct_mri}-(ii) shows the input MRI scan with artefacts and ground-truth.

\textbf{Training details and evaluation metrics.} All GANs are first initialized using the aforementioned progressive learning scheme with $(\lambda_1, \lambda_2)$ in Eq.~\ref{eq:eq2} set to $(1, 0.001)$.
We use Adam~\cite{kingma2014adam}, with the hyper-parameters
$\beta_1 := 0.9$, $\beta_2 := 0.999$, an initial learning rate of 0.002 for initialization and 0.0005 post-initialization that decays based on cosine
annealing over 1000 epochs, using a batch size of 8. 
We use three widely adopted metrics to evaluate image generation quality: 
PSNR measures $20 \log{\text{MAX}_I/\sqrt{\text{MSE}}}$, where $\text{MAX}_I$ is the highest possible intensity value in the image and $\text{MSE}$ is the mean-squared-error between two images.
SSIM computes the structural similarity between two images~\cite{wang2004image}. MAE computes the mean absolute error between two images. Higher PSNR, SSIM, and lower MAE indicate a higher quality of the generated images (wrt ground-truth). 

\textbf{Compared methods.} 
We compare our model to 
representative state-of-the-art methods for medical image translation, including
Pix2pix~\cite{pix2pix}, a baseline conditional adversarial networks for image-to-image translation tasks using GANs, PAN~\cite{wang2018perceptual}, and MedGAN~\cite{medgan}, a GAN-based method that relies on \textit{external-pre-trained feature extractors}, with a generator that refines the generated images progressively. MedGAN is shown to perform superior to methods like, Fila-sGAN~\cite{zhao2017synthesizing}, ID-cGAN~\cite{zhang2019image}, and achieve state-of-the-art performance for several medical image-to-image translation problems.

{
\setlength{\tabcolsep}{1.6pt}
\renewcommand{\arraystretch}{1.7}
\begin{table*}[t]
\resizebox{\textwidth}{!}
{
\begin{tabular}{l|ccc|ccc|ccc}
%\hline
    \multirow{2}{*}{\textbf{Methods}} 
  & \multicolumn{3}{c|}{\textbf{PET to CT}} 
  & \multicolumn{3}{c|}{\textbf{Undersampled MRI Recon.}} 
  & \multicolumn{3}{c}{\textbf{MRI Motion Correction}} \\ 
  & SSIM & PSNR & MAE & SSIM & PSNR & MAE & SSIM & PSNR & MAE \\
    \hline
  pix2pix~\cite{pix2pix} & 0.89$\pm$0.04 & 26.0$\pm$2.0 & 38.5$\pm$10.7 & 0.92$\pm$0.03 & 28.5$\pm$0.9 & 27.6$\pm$9.3 & 0.94$\pm$0.06 & 29.6$\pm$1.4 & 26.3$\pm$8.2 \\
  PAN~\cite{wang2018perceptual} & 0.90$\pm$0.08 & 26.5$\pm$4.5 & 37.2$\pm$15.6 & 0.93$\pm$0.05 & 28.8$\pm$0.7 & 26.2$\pm$10.4 & 0.95$\pm$0.10 & 30.1$\pm$2.8 & 24.9$\pm$9.7 \\
    %\hline
  MedGAN~\cite{medgan} & 0.90$\pm$0.04 & 27.1$\pm$2.5 & 35.4$\pm$11.8 & 0.94$\pm$0.02 & \textbf{29.7$\pm$1.9} & 24.2$\pm$8.7 & 0.95$\pm$0.04 & 30.8$\pm$1.8 & 23.6$\pm$9.1 \\
    \hline
%   Ours 
  \bf UP-GAN & \textbf{0.95$\pm$0.05} & \textbf{28.9$\pm$0.4} & \textbf{24.7$\pm$12.9} & \textbf{0.97$\pm$0.07} & 29.4$\pm$2.1 & \textbf{24.1$\pm$7.5} & \textbf{0.96$\pm$0.03} & \textbf{32.1$\pm$0.3} & \textbf{22.8$\pm$11.1} \\
\end{tabular}
}
\caption{
Evaluation of various methods on three medical image translation tasks. 
}
\vspace{-12pt}
\label{tab:quant}
\end{table*}
}

%\vspace{-10pt}
\subsection{Results and Analysis}
\label{sec:results}

\textbf{Qualitative results.} 
Figure~\ref{fig:uncerprog} visualizes the (intermediate) outputs of the generators at different phases of the framework. The visual quality of the generated image content increasingly improves along the network phases (as shown in the first column, second row onward). At the same time, prediction error and uncertainty decrease continuously (second column and fifth column, second row onward, respectively). High uncertainty values are found in anatomical regions with fine osseous structures, such as the nasal cavity and the inner ear in the petrous portion of the temporal bone. Particularly in such regions of high uncertainty, we achieve a progressive improvement in the level of detail.

Figure~\ref{fig:qual_petct_mri}-(Top) visualizes the generated CT images from the PET for all the compared methods along with our methods. We observe that more high-frequency features are present in our prediction compared to the previous state-of-the-art model (MedGAN). We also observe that the overall residual is significantly lower for our method compared to the other baselines. MedGAN performs better than pix2pix in synthesizing high-frequency features and sharper images. Figure~\ref{fig:qual_petct_mri}-(Bottom) shows similar results for the undersampled MRI reconstruction task and MRI motion correction task. In both cases, our model yields superior images, as can be seen via relatively neutral residual maps.

\textbf{Quantitative results.} 
Table~\ref{tab:quant} shows the quantitative performance of all the methods on the three tasks; for all the tasks, our method outperforms the recent models. In particular, for the most challenging task, PET to CT translation, our method with uncertainty-based guide outperforms the previous state-of-the-art method, MedGAN (that relies on task-specific external feature extractor),
\textit{without using any external feature extractor}. Therefore, the uncertainty guidance reduces the burden of having an externally trained task-specific feature extractor to achieve high fidelity images. The same trend holds for undersampled MRI reconstruction and motion correction in MRI.
The statistical tests on SSIM values of MedGAN and our UP-GAN gives us a \textit{p-value} of 0.016 for PET-to-CT translation, 0.021 for undersampled
MRI reconstruction, and 0.036 for MRI motion correction. 
As all the \textit{p-values} are $<0.05$, results are statistically
significant.

\textbf{Ablation study.} We study the model that does not utilize the estimated uncertainty maps as attention maps and observe that the model without the uncertainty as the guide performs inferior to the UP-GAN with a performance (SSIM/PSNR/MAE) of
(0.87/25.4/40.7), (0.93/27.3/38.7), and (0.92/26.2/35.1) for PET to CT translation, undersampled MRI reconstruction, and MRI motion correction, respectively. 
UP-GAN model leverages the uncertainty map to refine the predictions where the model is uncertain, which is also correlated to the regions where the translation is poor. The model without uncertainty-based guidance does not focus on the regions mentioned above in the prediction and is unable to perform as well as UP-GAN. 
\begin{figure}[!t]
\centering
\includegraphics[width=0.98\textwidth]{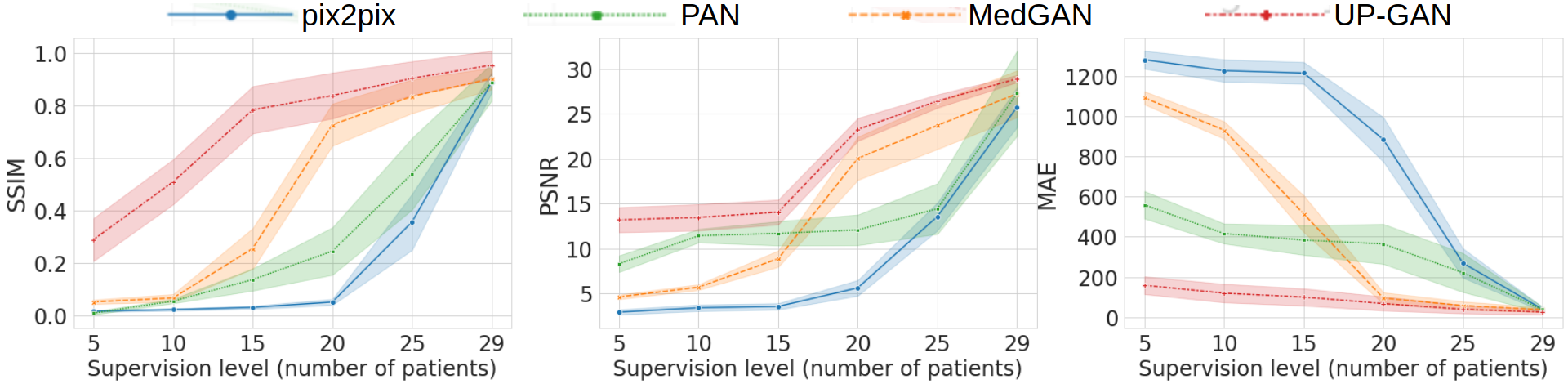}
\caption{
Quantitative results in the presence of limited labeled training data. 
} 
\label{fig:sl_graph}
\vspace{-1em}
\end{figure}

\textbf{Evaluating models with weak supervision.}
We evaluate all the models for PET to CT synthesis by limiting the number of paired image samples used for training. 
We define \textit{five} supervision levels corresponding to different amounts of cross-domain pairwise training sample slices. For this experiment, we train the recent state-of-the-art models with a varying number of patients in the training stage, i.e., we use 5, 10, 15, 20, and 29 patients, respectively. Figure~\ref{fig:sl_graph} shows the performance of all the models at varying supervision levels. We observe that our model with uncertainty guidance outperforms all the baselines at full supervision (with 29 patients). Moreover, our model sharply outperforms the baselines with limited training data (with $<$ 29 patients). UP-GAN produces intermediate uncertainty maps that have higher values under weak supervision (compared to the full supervision case), but this still allows UP-GAN to focus on highly uncertain regions, that the current state-of-the-art models do not have access to, hence are not able to leverage that to refine the predicted images.

%\vspace{-10pt}
\section{Conclusion}
%\vspace{-12pt}
In this work, we propose a new generic model for medical image translation using uncertainty-guided progressive GANs. We demonstrate how uncertainty can serve as an attention map in progressive learning schemes. We demonstrate the efficacy of our method on three challenging medical image translation tasks, including PET to CT translation, undersampled MRI reconstruction, and motion correction in MRI.  Our method achieves state-of-the-art in various tasks.
Moreover, it allows the quantification of uncertainty and shows better generalizability with smaller sample sizes than recent approaches.

\vskip 0.5em 
% \section*{Acknowledgements} 
\myparagraph{Acknowledgements.} This work has been partially funded by the ERC (853489 - DEXIM) and by the DFG (2064/1 – Project number 390727645). The authors thank the International Max Planck Research School for Intelligent Systems (IMPRS-IS)
for support.

%
% ---- Bibliography ----
%
% BibTeX users should specify bibliography style 'splncs04'.
% References will then be sorted and formatted in the correct style.
%
\bibliographystyle{splncs04}
\bibliography{paper}

\newpage
\section*{Appendix}
The following presents the numerical results for the weak supervision evaluation.

{
\setlength{\tabcolsep}{1.6pt}
\renewcommand{\arraystretch}{1}
\begin{table*}
\resizebox{\textwidth}{!}
{
\begin{tabular}{l|cccc}
%\hline
    \multirow{2}{*}{\textbf{Methods}} 
  & \multicolumn{4}{c}{\textbf{PET to CT}} \\ 
  & SSIM & PSNR & MAE & log(MAE) \\
    \hline
  pix2pix~\cite{pix2pix} & 0.02$\pm$0.003 & 2.91$\pm$0.29 & 1281$\pm$46.94 & 7.15$\pm$0.035 \\
  PAN~\cite{wang2018perceptual} & 0.01$\pm$0.006 & 8.3$\pm$0.93 & 557$\pm$68.6 & 6.31$\pm$0.12 \\
    %\hline
  MedGAN~\cite{medgan} & 0.05$\pm$0.008 & 4.6$\pm$0.25 & 1090$\pm$35.7 & 6.99$\pm$0.03 \\
    \hline
%   Ours 
  \bf UP-GAN & \textbf{0.29$\pm$0.08} & \textbf{13.2$\pm$1.39} & \textbf{158.1$\pm$43.41} & \textbf{5.02$\pm$0.32} \\
\end{tabular}
}
\caption{
Evaluation of different models trained with data from 5 patients.
}
\vspace{-12pt}
\label{tab:quant}
\end{table*}
}

{
\setlength{\tabcolsep}{1.6pt}
\renewcommand{\arraystretch}{1}
\begin{table*}
\resizebox{\textwidth}{!}
{
\begin{tabular}{l|cccc}
%\hline
    \multirow{2}{*}{\textbf{Methods}} 
  & \multicolumn{4}{c}{\textbf{PET to CT}} \\ 
  & SSIM & PSNR & MAE & log(MAE) \\
    \hline
  pix2pix~\cite{pix2pix} & 0.03$\pm$0.005 & 3.55$\pm$0.36 & 1214$\pm$53.98 & 7.10$\pm$0.044 \\
  PAN~\cite{wang2018perceptual} & 0.14$\pm$0.04 & 11.68$\pm$1.36 & 383$\pm$78.4 & 5.94$\pm$0.20 \\
    %\hline
  MedGAN~\cite{medgan} & 0.26$\pm$0.08 & 8.86$\pm$0.85 & 512$\pm$88.8 & 6.22$\pm$0.18 \\
    \hline
%   Ours 
  \bf UP-GAN & \textbf{0.78$\pm$0.09} & \textbf{14.09$\pm$1.41} & \textbf{98.9$\pm$42.64} & \textbf{4.48$\pm$0.55} \\
\end{tabular}
}
\caption{
Evaluation of different models trained with data from 15 patients.
}
\vspace{-12pt}
\label{tab:quant}
\end{table*}
}

{
\setlength{\tabcolsep}{1.6pt}
\renewcommand{\arraystretch}{1}
\begin{table*}
\resizebox{\textwidth}{!}
{
\begin{tabular}{l|cccc}
%\hline
    \multirow{2}{*}{\textbf{Methods}} 
  & \multicolumn{4}{c}{\textbf{PET to CT}} \\ 
  & SSIM & PSNR & MAE & log(MAE) \\
    \hline
  pix2pix~\cite{pix2pix} & 0.36$\pm$0.11 & 13.56$\pm$1.53 & 269$\pm$72.54 & 5.56$\pm$0.27 \\
  PAN~\cite{wang2018perceptual} & 0.54$\pm$0.13 & 14.38$\pm$2.70 & 221$\pm$90.4 & 5.30$\pm$0.44 \\
    %\hline
  MedGAN~\cite{medgan} & 0.83$\pm$0.07 & 23.65$\pm$2.65 & 57.48$\pm$20.72 & 3.99$\pm$0.33 \\
    \hline
%   Ours 
  \bf UP-GAN & \textbf{0.90$\pm$0.07} & \textbf{26.40$\pm$0.74} & \textbf{38.9$\pm$20.69} & \textbf{3.49$\pm$0.62} \\
\end{tabular}
}
\caption{
Evaluation of different models trained with data from 25 patients.
}
\vspace{-12pt}
\label{tab:quant}
\end{table*}
}

\end{document}